\RequirePackage[svgnames]{xcolor}

\documentclass[10pt, a4paper]{article}

\usepackage[final]{lrec2026} 
\usepackage{comment}
\usepackage{booktabs, multirow} 
\usepackage{soul} 
\usepackage{xcolor,colortbl} 
\usepackage{changepage,threeparttable} 
\usepackage{soul}
\usepackage{amsmath}
\usepackage[capitalise,nameinlink]{cleveref}
\usepackage{tcolorbox}
\usepackage{wrapfig}
\usepackage{hyperref}

\newcommand{\rowhighlight}{\cellcolor[gray]{0.9}}

\title{Trust but Verify: Introducing DAVinCI - A Framework for \ul{D}ual \ul{A}ttribution and \ul{V}erification in \ul{C}laim \ul{I}nference for Language Models}
\name{Vipula Rawte\textsuperscript{1}, Ryan Rossi\textsuperscript{2}, Franck Dernoncourt\textsuperscript{2}, Nedim Lipka\textsuperscript{2}} 

\address{\textsuperscript{1}Adobe, \textsuperscript{2}Adobe Research \\
         vrawte@adobe.com\\}

\abstract{
Large Language Models (LLMs) have demonstrated remarkable fluency and versatility across a wide range of NLP tasks, yet they remain prone to factual inaccuracies and hallucinations. This limitation poses significant risks in high-stakes domains such as healthcare, law, and scientific communication, where trust and verifiability are paramount. In this paper, we introduce DAVinCI - a Dual Attribution and Verification framework designed to enhance the factual reliability and interpretability of LLM outputs. DAVinCI operates in two stages: (i) it attributes generated claims to internal model components and external sources; (ii) it verifies each claim using entailment-based reasoning and confidence calibration. We evaluate DAVinCI across multiple datasets, including FEVER and CLIMATE-FEVER, and compare its performance against standard verification-only baselines. Our results show that DAVinCI significantly improves classification accuracy, attribution precision, recall, and F1-score by \textbf{5-20\%}. Through an extensive ablation study, we isolate the contributions of evidence span selection, recalibration thresholds, and retrieval quality. We also release a modular DAVinCI implementation that can be integrated into existing LLM pipelines. By bridging attribution and verification, DAVinCI offers a scalable path to auditable, trustworthy AI systems. This work contributes to the growing effort to make LLMs not only powerful but also accountable. Our code is available at \url{https://github.com/vr25/davinci}.
\\ \newline  }

\begin{document}

\maketitleabstract

\section{Introduction}

The rapid proliferation of large language models (LLMs) has revolutionized natural language generation, enabling systems to produce fluent and contextually rich text across diverse domains. However, this fluency often masks a critical vulnerability: the tendency of LLMs to generate factually incorrect or hallucinated content \cite{rawte-etal-2023-troubling,rawte2023survey,liu2024surveyhallucinationlargevisionlanguage}. As LLMs become increasingly embedded in high-stakes applications - such as scientific writing, journalism, and legal analysis - the need for robust factuality verification has become paramount \cite{thorne-vlachos-2018-automated,jaradat-etal-2018-claimrank,augenstein-etal-2019-multifc,10.1007/978-3-030-15719-7_41,dmonte2025claimverificationagelarge}.

Recent studies have highlighted the limitations of LLMs in maintaining factual consistency, especially when responding to open-ended prompts or synthesizing information from multiple sources \cite{rahman2025hallucinationtruthreviewfactchecking}. To address this, researchers have explored various verification pipelines, including retrieval-augmented generation (RAG) \cite{lewis2021retrievalaugmentedgenerationknowledgeintensivenlp}, entailment-based classifiers, and attribution-aware filtering. Yet, most existing systems treat attribution and verification as isolated components, failing to capture the nuanced interplay between evidence selection and entailment reasoning.

To address these limitations, we propose \textbf{DAVinCI} - a Dual Attribution and Verification framework for trustworthy LLMs. DAVinCI operates in two stages: (i) it attributes claims to internal model components and external sources; (ii) it verifies each claim using entailment-based reasoning and confidence calibration. This dual approach enables LLMs to not only generate claims but also justify and validate them in a transparent and auditable manner.

This approach builds on insights from recent benchmarks such as FactBench \cite{bayat2025factbenchdynamicbenchmarkinthewild}, which categorize hallucination-prone prompts and evaluate factuality across difficulty tiers. It also aligns with broader efforts to characterize and correct factual errors in LLM outputs. Moreover, DAVinCI’s modular design allows researchers to isolate and evaluate the impact of attribution quality, verification robustness, and calibration sensitivity - three pillars increasingly recognized as essential for trustworthy AI systems.

In this paper, we present a comprehensive evaluation of DAVinCI across multiple entailment models and attribution configurations. We demonstrate that full-passage evidence consistently improves verification accuracy and F1-score, while confidence recalibration enables flexible trade-offs between precision and recall.

Our main \textbf{contributions} are as follows:

\begin{itemize}
    
    \item DAVinCI, a modular framework that integrates attribution and verification into LLM inference.

    \item We conduct an ablation study to isolate the impact of evidence span selection, recalibration thresholds, and retrieval quality.

    \item We release a reproducible implementation of DAVinCI to support future research in trustworthy NLP.
    
\end{itemize}

\section{Related Work}

The challenge of ensuring factual correctness in natural language generation has led to a growing body of research in claim verification, retrieval-augmented generation, and model attribution. Early work in automated fact-checking focused on datasets such as FEVER \cite{thorne2018fever}, which introduced the task of verifying claims against evidence retrieved from Wikipedia. Subsequent efforts like CLIMATE-FEVER \cite{Diggelmann2020CLIMATEFEVERAD} extended this paradigm to other domains, emphasizing the need for domain-specific reasoning and evidence selection.

Retrieval-Augmented Generation (RAG) frameworks \cite{lewis2021retrievalaugmentedgenerationknowledgeintensivenlp} have emerged as a popular solution for grounding LLM outputs in external sources. These models combine dense retrievers with generative architectures to produce fluent, evidence-aware responses. However, most RAG systems treat retrieval and generation as loosely coupled stages, without explicit verification of factual consistency. Recent work has attempted to address this gap through hybrid models that incorporate entailment classifiers or post-hoc verification modules \cite{shuster-etal-2022-language}.

Attribution in LLMs remains an underexplored area. While Attributable QA \cite{liu-etal-2023-evaluating} have begun to formalize attribution as a structured task, but these approaches are often limited to specific domains or require extensive supervision.

Confidence calibration has also gained traction as a means of improving trust in model predictions. Techniques such as temperature scaling \cite{Guo2017OnCO} and trust-aware scoring \cite{10.1145/3465481.3470093} aim to align model confidence with prediction reliability. In the context of claim verification, calibrated confidence can help distinguish between supported, refuted, and uncertain claims - a capability that is central to our DAVinCI framework.

Our work builds on these foundations by integrating attribution and verification into a unified pipeline. Unlike prior approaches that treat these components independently, DAVinCI leverages attribution to inform verification, and uses calibrated confidence to produce interpretable and trustworthy outputs. To our knowledge, this is the first framework to combine internal and external attribution with entailment-based verification in a modular and scalable way.

\section{Proposed Method}

The DAVinCI framework is designed to enhance the factual reliability of LLM outputs by integrating two complementary stages: Attribution and Verification. This section details the architecture, components, and workflow of DAVinCI, along with the rationale behind key design choices.

\begin{figure*}[!]
    \centering
    \includegraphics[width=\linewidth]{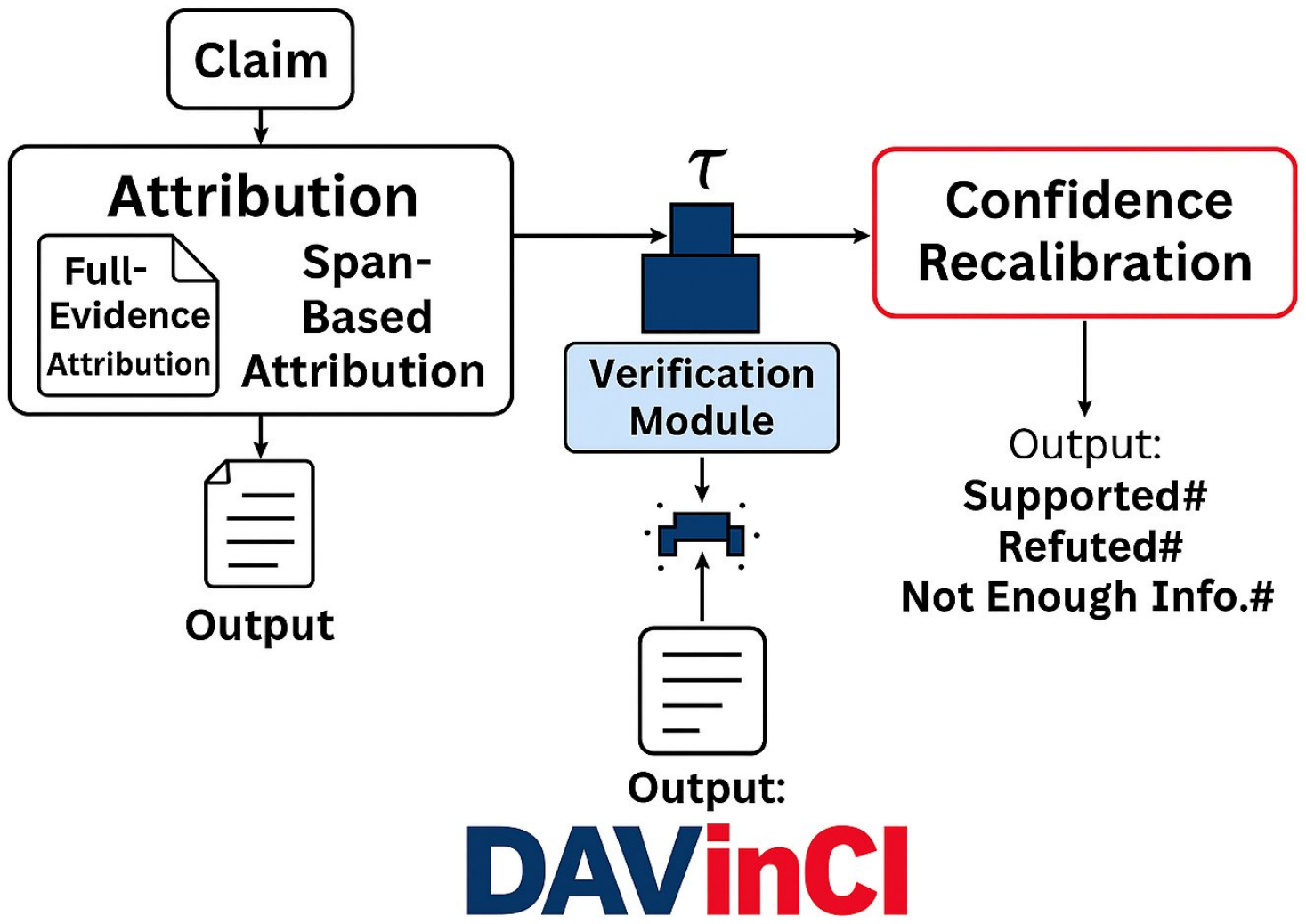}
    \caption{\textbf{Claim Input:} The process begins with a textual claim, typically generated by a language model or extracted from a corpus. This claim serves as the input to the attribution module. \textbf{Attribution Module:} This module identifies relevant evidence passages that support or refute the claim. It operates in two modes: \textbf{(a) Full-Evidence Attribution} retrieves entire passages based on semantic similarity to the claim. \textbf{(b) Span-Based Attribution} extracts specific answer spans using a question-answering model, offering fine-grained attribution. These two strategies provide complementary views of evidence relevance. \textbf{Evidence Output:} The selected evidence is passed downstream for verification. This intermediate output may include gold-standard evidence (from annotated datasets) or retrieved evidence (from a corpus). \textbf{Verification Module:} This component uses a transformer-based entailment classifier to assess the relationship between the claim and its attributed evidence. It outputs a label from the set {Supported, Refuted, Not Enough Info} along with a confidence score s$\in$[0,1]. \textbf{Confidence Recalibration:} To mitigate overconfident misclassifications, the confidence score is passed through a thresholding mechanism. If the score s falls below a predefined threshold $\tau$, the output label is downgraded to Not Enough Info\#, regardless of the original prediction. This ensures conservative decision-making in ambiguous cases. \textbf{Final Output:} The system produces a calibrated label - Supported\#, Refuted\#, or Not Enough Info\# - which reflects both the entailment judgment and the confidence adjustment. These outputs can be used for downstream tasks such as fact-checking, claim filtering, or human-in-the-loop review.}
    
    \label{fig:placeholder}
\end{figure*}

DAVinCI operates in a two-stage pipeline:

1. \textbf{Attribution:} Given a claim, the framework identifies relevant evidence from internal or external sources that could plausibly support or refute the claim.

2. \textbf{Verification:} The attributed evidence is then passed to an entailment-based verifier that classifies the claim as Supported, Refuted, or Not Enough Information (NEI), optionally applying confidence recalibration.

This dual structure allows DAVinCI to not only assess the truthfulness of a claim but also provide interpretable justifications grounded in evidence. Here is an example from the FEVER dataset showing a claim, the evidence behind it, and its label.

\begin{tcolorbox}[colframe=red, colback=red!10, coltitle=black, top=2.5mm, bottom=2.5mm, left=1.5mm, right=1.5mm]

\scriptsize \ttfamily \textbf{CLAIM:} Caroline Kennedy is American.
\end{tcolorbox}
\vspace{-5mm}
\begin{tcolorbox}[colframe=green, colback=green!10, coltitle=black, top=1.5mm, bottom=1.5mm, left=1.5mm, right=1.5mm]

\scriptsize \ttfamily \textbf{EVIDENCE:} Caroline Bouvier Kennedy (born November 27, 1957) is an American author, attorney, and diplomat who served as the United States Ambassador to Japan from 2013 to 2017.

\end{tcolorbox}
\vspace{-5mm}
\begin{tcolorbox}[colframe=cyan, colback=cyan!10, coltitle=black, top=2.5mm, bottom=2.5mm, left=1.5mm, right=1.5mm]

\scriptsize \ttfamily \textbf{LABEL:} ENTAILMENT

\end{tcolorbox}

Let \( c \) be a claim generated by a language model, and let \( \mathcal{E} = \{e_1, e_2, \dots, e_n\} \) be a set of retrieved evidence passages.

\subsection{Attribution Module}

The attribution module is responsible for identifying the most relevant evidence for a given claim. We explore two strategies:

\textbf{Full Evidence Attribution:} In datasets like FEVER and CLIMATE-FEVER, gold-standard evidence is available. We use the full evidence text directly without modification.

\textbf{Span-Based Attribution:} We simulate a retrieval scenario by using a question-answering (QA) model (\texttt{deepset/roberta-base-squad2-distilled}\footnote{\url{https://huggingface.co/deepset/roberta-base-squad2-distilled}}) to extract the most relevant span from the evidence given the claim as a question. This mimics real-world settings where only partial evidence may be accessible.

This module can be extended to incorporate dense retrievers (e.g., DPR, E5) or hybrid retrieval systems in future work.

\subsubsection{Attribution Scoring}

Given a claim \( c \) and a set of evidence passages \( \{e_1, e_2, \dots, e_n\} \), we compute attribution scores using either full-evidence or span-based retrieval:
\[
\text{Score}_{\text{attr}}(c, e_i) = \text{sim}(c, e_i)
\]
where \( \text{sim}(\cdot) \) is a similarity function such as cosine similarity or dense retriever score.

\subsection{Verification Module}

The verification module takes the claim and attributed evidence as input and outputs a label from {Supported, Refuted, Not Enough Information}. We use a transformer-based entailment classifier model trained on multiple NLI datasets.

\begin{tcolorbox}
[boxsep=0pt,left=2.5pt,right=5pt,top=3pt,bottom=3pt,colback=Wheat!55!white,colframe=Wheat!45!black]
\scriptsize
\textbf{Input Format}  \\
\texttt{[Claim] [SEP] [Attributed Evidence]}
\end{tcolorbox}

The model outputs a label and a confidence score. This score is used in downstream calibration.

Each claim-evidence pair \( (c, e_i) \) is passed to an entailment classifier \( f_{\text{ver}} \) that outputs a label \( y \in \{\text{Supported}, \text{Refuted}, \text{Not Enough Info}\} \) and a confidence score \( s \in [0, 1] \):
\[
(y, s) = f_{\text{ver}}(c, e_i)
\]

\subsection{Confidence Recalibration}

To mitigate overconfidence and improve trustworthiness, we apply a simple threshold-based recalibration. If the model’s confidence score is below a threshold \( \tau \) (default: 0.6), the prediction is overridden to NEI. This reflects epistemic uncertainty and aligns with human fact-checking behavior.

To reduce overconfident misclassifications, we apply a threshold \( \tau \) to the confidence score \( s \). The final output label \( y^\# \) is defined as:
\[
y^\# =
\begin{cases}
y & \text{if } s \geq \tau \\
\text{Not Enough Info}^\# & \text{otherwise}
\end{cases}
\]

\paragraph{Final Decision}

For multiple evidence passages, we aggregate verification scores and labels using majority voting or weighted averaging:
\[
y_{\text{final}} = \text{Aggregate}(\{y_1^\#, y_2^\#, \dots, y_n^\#\})
\]

\section{Experiments} 

\subsection{Dataset}

We assess DAVinCI using two prominent datasets for claim verification: FEVER\footnote{\url{https://huggingface.co/datasets/tommasobonomo/sem\_augmented\_fever\_nli}} and CLIMATE-FEVER\footnote{\url{https://huggingface.co/datasets/rexarski/climate\_fever\_fixed}}. Both datasets offer annotated claims, supporting evidence, and ground-truth labels, making them well-suited for evaluating attribution and verification workflows.

\textbf{FEVER:} A curated subset of the original FEVER dataset, featuring claims categorized as contradiction, entailment, or neutral, accompanied by verified evidence sentences sourced from Wikipedia.

\textbf{CLIMATE-FEVER:} A domain-specific fact-checking dataset focused on climate change. It includes claims extracted from scientific literature, paired with abstracts and rationale spans. Labels are assigned as Supported, Refuted, or Not Enough Information (NEI). The accompanying table outlines dataset statistics and class distribution.

Both datasets provide gold-standard evidence, enabling the evaluation of both full-evidence and span-level attribution techniques. \cref{tab:data-stat} shows the datatset distribution.

\begin{table}[!]
\scriptsize \centering
\begin{tabular}{l|rrr}\toprule
\textbf{Class label } &\textbf{FEVER} &\textbf{CLIMATE-FEVER} \\\midrule
\textbf{ENTAILMENT / SUPPORTS} &792 &375 \\
\textbf{CONTRADICTION / REFUTES} &812 &164 \\
\textbf{NEUTRAL / NOT\_ENOUGH\_INFO} &683 &996 \\ \midrule
\textbf{Total} &\textbf{2287} &\textbf{1535} \\
\bottomrule
\end{tabular}
\caption{Class label distribution in the FEVER and CLIMATE-FEVER datasets}\label{tab:data-stat}
\end{table}

\subsection{Baselines}

To assess the effectiveness of our proposed approach, we conducted a comprehensive evaluation against several state-of-the-art Natural Language Inference (NLI) models. Specifically, we benchmarked our method against the following models which are widely recognized for their strong performance in language understanding tasks. 

\begin{enumerate}

    \item \texttt{microsoft/deberta-large-mnli}\footnote{\url{https://huggingface.co/microsoft/deberta-large-mnli}}

    \item \texttt{FacebookAI/roberta-large-mnli}\footnote{\url{https://huggingface.co/FacebookAI/roberta-large-mnli}}

    \item \texttt{facebook/bart-large-mnli}\footnote{\url{https://huggingface.co/facebook/bart-large-mnli}}

    \item \texttt{ynie/roberta-large-snli\_mnli\_fever\_anli\_R1\_R2\_R3-nli}\footnote{\url{https://huggingface.co/ynie/roberta-large-snli\_mnli\_fever\_anli\_R1\_R2\_R3-nli}}
    
\end{enumerate}

The comparison was carried out on the verification entailment task, allowing us to measure how well each model identifies and validates factual relationships within textual data.

We compare DAVinCI against a standard verification-only baseline:

\begin{itemize}
    \item \textbf{Baseline:} Verifier-only using full evidence.
    \item \textbf{DAVinCI-Recalibrated:} QA span + verification + recalibration (with $\tau$=0.6).
\end{itemize}


This setup allows us to isolate the impact of attribution quality and confidence calibration.

\subsection{Implementation Details}

All experiments were performed on an Apple MacBook equipped with the M4 chip, 32GB of RAM, and a 10-core CPU - offering a robust and energy-efficient setup optimized for high-performance computing. We used the Hugging Face Transformers library\footnote{\url{https://huggingface.co}} for all models. We deliberately selected models and datasets under the \texttt{MIT license} to ensure safer and more straightforward reproducibility.


\begin{table*}[!]
\footnotesize \centering
\begin{tabular}{l|c|r|rrrrr}\toprule
\textbf{} &\textbf{Model} &\textbf{} &\textbf{Precision ↑} &\textbf{Recall ↑} &\textbf{F1-Score ↑} &\textbf{Accuracy ↑} \\\midrule
\multirow{12}{*}{\rotatebox{90}{\textbf{Baseline}}} &\multirow{3}{*}{\textbf{deberta-large-mnli}} &\textbf{} & \rowhighlight & \rowhighlight & \rowhighlight & \rowhighlight 0.42 \\
& &\textbf{macro avg} & \rowhighlight 0.52 & \rowhighlight 0.44 & \rowhighlight 0.36 & \rowhighlight \\
& &\textbf{weighted avg} & \rowhighlight 0.53 & \rowhighlight 0.42 & \rowhighlight 0.36 & \rowhighlight \\
&\multirow{3}{*}{\textbf{roberta-large-mnli}} &\textbf{} & & & &0.36 \\
& &\textbf{macro avg} &0.43 &0.38 &0.3 & \\
& &\textbf{weighted avg} &0.43 &0.36 &0.29 & \\
&\multirow{3}{*}{\textbf{bart-large-mnli}} &\textbf{} & \rowhighlight & \rowhighlight & \rowhighlight & \rowhighlight 0.42 \\
& &\textbf{macro avg} & \rowhighlight 0.46 & \rowhighlight 0.43 & \rowhighlight 0.36 & \rowhighlight \\
& &\textbf{weighted avg} & \rowhighlight 0.46 & \rowhighlight 0.42 & \rowhighlight 0.36 & \rowhighlight \\
&\multirow{3}{*}{\textbf{roberta-large-snli}} &\textbf{} & & & &0.38 \\
& &\textbf{macro avg} &0.46 &0.4 &0.34 & \\
& &\textbf{weighted avg} &0.46 &0.38 &0.34 & \\ \midrule
\multirow{12}{*}{\rotatebox{90}{\textbf{DAVinCI-Recalibrated}}} &\multirow{3}{*}{\textbf{deberta-large-mnli}} &\textbf{} & \rowhighlight \textbf{} & \rowhighlight \textbf{} & \rowhighlight \textbf{} & \rowhighlight \textbf{0.48} \\
& &\textbf{macro avg} & \rowhighlight \textbf{0.61} & \rowhighlight \textbf{0.49} & \rowhighlight \textbf{0.41} & \rowhighlight \textbf{} \\
& &\textbf{weighted avg} & \rowhighlight \textbf{0.62} & \rowhighlight \textbf{0.48} & \rowhighlight \textbf{0.41} & \rowhighlight \textbf{} \\
&\multirow{3}{*}{\textbf{roberta-large-mnli}} &\textbf{} &\textbf{} &\textbf{} &\textbf{} &\textbf{0.44} \\
& &\textbf{macro avg} &\textbf{0.51} &\textbf{0.45} &\textbf{0.38} &\textbf{} \\
& &\textbf{weighted avg} &\textbf{0.52} &\textbf{0.44} &\textbf{0.38} &\textbf{} \\
&\multirow{3}{*}{\textbf{bart-large-mnli}} &\textbf{} & \rowhighlight \textbf{} & \rowhighlight \textbf{} & \rowhighlight \textbf{} & \rowhighlight \textbf{0.43} \\
& &\textbf{macro avg} & \rowhighlight \textbf{0.51} & \rowhighlight \textbf{0.44} & \rowhighlight \textbf{0.37} & \rowhighlight \textbf{} \\
& &\textbf{weighted avg} & \rowhighlight \textbf{0.52} & \rowhighlight \textbf{0.43} & \rowhighlight \textbf{0.37} & \rowhighlight \textbf{} \\
&\multirow{3}{*}{\textbf{roberta-large-snli}} &\textbf{} &\textbf{} &\textbf{} &\textbf{} &\textbf{0.42} \\
& &\textbf{macro avg} &\textbf{0.57} &\textbf{0.44} &\textbf{0.4} &\textbf{} \\
& &\textbf{weighted avg} &\textbf{0.59} &\textbf{0.42} &\textbf{0.4} &\textbf{} \\
\bottomrule
\end{tabular}
\caption{\textbf{FEVER dataset:} Evaluation of DAVinCI's performance compared to baseline verification, including macro and weighted averages for F1 score, precision, and recall across three classes: entailment, contradiction, and neutral. The best-performing results are highlighted in bold.}\label{tab:fever-res}
\end{table*}


\begin{table*}[!]
\footnotesize \centering
\begin{tabular}{l|c|r|rrrrr}\toprule
\textbf{} &\textbf{Model} &\textbf{} &\textbf{Precision ↑} &\textbf{Recall ↑} &\textbf{F1-Score ↑} &\textbf{Accuracy ↑} \\\midrule
\multirow{12}{*}{\rotatebox{90}{\textbf{Baseline}}} &\multirow{3}{*}{\textbf{deberta-large-mnli}} &\textbf{} & \rowhighlight & \rowhighlight & \rowhighlight & \rowhighlight 0.6 \\
& &\textbf{macro avg} & \rowhighlight 0.41 & \rowhighlight 0.38 & \rowhighlight 0.34 & \rowhighlight \\
& &\textbf{weighted avg} & \rowhighlight 0.53 & \rowhighlight 0.6 & \rowhighlight 0.51 & \rowhighlight \\
&\multirow{3}{*}{\textbf{roberta-large-mnli}} &\textbf{} & & & &0.6 \\
& &\textbf{macro avg} &0.48 &0.41 &0.38 & \\
& &\textbf{weighted avg} &0.58 &0.6 &0.54 & \\
&\multirow{3}{*}{\textbf{bart-large-mnli}} &\textbf{} & \rowhighlight & \rowhighlight & \rowhighlight & \rowhighlight 0.58 \\
& &\textbf{macro avg} & \rowhighlight 0.52 & \rowhighlight 0.43 & \rowhighlight 0.38 & \rowhighlight \\
& &\textbf{weighted avg} & \rowhighlight 0.61 & \rowhighlight 0.58 & \rowhighlight 0.52 & \rowhighlight \\
&\multirow{3}{*}{\textbf{roberta-large-snli}} &\textbf{} & & & &0.65 \\
& &\textbf{macro avg} &0.54 &0.35 &0.31 & \\
& &\textbf{weighted avg} &0.62 &0.65 &0.54 & \\ \midrule
\multirow{12}{*}{\rotatebox{90}{\textbf{DAVinCI-Recalibrated}}} &\multirow{3}{*}{\textbf{deberta-large-mnli}} &\textbf{} & \rowhighlight & \rowhighlight & \rowhighlight  & \rowhighlight \textbf{0.63} \\
& &\textbf{macro avg} & \rowhighlight \textbf{0.55} & \rowhighlight \textbf{0.44} & \rowhighlight \textbf{0.4} & \rowhighlight \textbf{} \\
& &\textbf{weighted avg} & \rowhighlight \textbf{0.62} & \rowhighlight \textbf{0.63} & \rowhighlight \textbf{0.55} & \rowhighlight \textbf{} \\
&\multirow{3}{*}{\textbf{roberta-large-mnli}} &\textbf{} &\textbf{} &\textbf{} &\textbf{} &\textbf{0.63} \\
& &\textbf{macro avg} &\textbf{0.56} &\textbf{0.46} &\textbf{0.44} &\textbf{} \\
& &\textbf{weighted avg} &\textbf{0.63} &\textbf{0.63} &\textbf{0.57} &\textbf{} \\
&\multirow{3}{*}{\textbf{bart-large-mnli}} &\textbf{} & \rowhighlight \textbf{} & \rowhighlight \textbf{} & \rowhighlight \textbf{} & \rowhighlight \textbf{0.6} \\
& &\textbf{macro avg} & \rowhighlight 0.52 & \rowhighlight 0.43 & \rowhighlight \textbf{0.39} & \rowhighlight \textbf{} \\
& &\textbf{weighted avg} & \rowhighlight 0.61 & \rowhighlight \textbf{0.6} & \rowhighlight \textbf{0.54} & \rowhighlight \textbf{} \\
&\multirow{3}{*}{\textbf{roberta-large-snli}} &\textbf{} &\textbf{} &\textbf{} &\textbf{} &\textbf{0.66} \\
& &\textbf{macro avg} &\textbf{0.61} &\textbf{0.39} &\textbf{0.38} &\textbf{} \\
& &\textbf{weighted avg} &\textbf{0.65} &\textbf{0.66} &\textbf{0.56} &\textbf{} \\
\bottomrule
\end{tabular}
\caption{\textbf{CLIMATE-FEVER dataset:} Comparison of DAVinCI's performance with baseline verification, reporting macro and weighted averages for F1 score, precision, and recall across the three classification categories: entailment, contradiction, and neutral. Accuracy is also provided for all four language models. Top-performing results are indicated in bold. }\label{tab:climate-res}
\end{table*}

\subsection{Main results}

\cref{tab:fever-res,tab:climate-res} show the main results for performance of DAVinCI on both FEVER and CLIMATE-FEVER datasets. The performance comparison between baseline and Davinci-enhanced models reveals consistent improvements across all metrics. For the \texttt{deberta-large-mnli model}, Davinci boosts accuracy from 0.42 to 0.48, with macro F1-score rising from 0.36 to 0.41. Similarly, the \texttt{roberta-large-mnli} model sees its accuracy increase from 0.36 to 0.44, and macro F1-score from 0.30 to 0.38. The \texttt{bart-large-mnli} model also benefits, with accuracy improving from 0.42 to 0.43 and macro F1-score from 0.36 to 0.37. Lastly, the \texttt{roberta-large-snli} model shows gains in accuracy (0.38 to 0.42) and macro F1-score (0.34 to 0.40).

Across all models, Davinci integration leads to higher precision, recall, and F1-scores in both macro and weighted averages. The \texttt{deberta-large-mnli} model under Davinci achieves the highest macro precision (0.61) and weighted precision (0.62), indicating strong overall classification performance. Even models with lower baseline scores, like \texttt{roberta-large-mnli}, show notable improvements under Davinci, suggesting that the enhancement consistently elevates model robustness and reliability. These results underscore Davinci's effectiveness in refining natural language inference tasks across diverse transformer architectures.
\section{Ablation Study}

We conduct an ablation study to evaluate the impact of individual components by comparing full evidence \cref{tab:fever-abl,tab:climate-abl}, span-based evidence, and varying threshold settings (0.7, 0.8, 0.9) \cref{tab:fever-thresh,tab:climate-thresh}, enabling us to measure how each design choice influences performance and trustworthiness.

\subsection{DAVinCI with \textit{full} evidence}

The full-evidence variant consistently outperforms the span-based counterpart across all metrics. Among four entailment models, \texttt{roberta-large-snli} leads with 0.48 accuracy, macro F1, and weighted F1, showing balanced class performance. \texttt{deberta-large-mnli} excels in macro precision (0.57) and weighted precision (0.58), indicating confident predictions. All full-evidence models maintain macro and weighted F1-scores above 0.37, with recall between 0.44 and 0.49, highlighting improved class balance and semantic grounding.
For CLIMATE-FEVER, full-evidence models again perform best. \texttt{roberta-large-snli} achieves 0.65 accuracy and 0.57 weighted F1. \texttt{deberta-large-mnli} and \texttt{roberta-large-mnli} follow with 0.61 accuracy and weighted F1 above 0.54. Macro precision ranges from 0.49 to 0.59, recall from 0.40 to 0.47, and weighted F1 consistently exceeds 0.54, confirming reliable and balanced predictions.

\subsection{DAVinCI with \textit{span} evidence}

Span-evidence models show weaker performance across the board. \texttt{bart-large-mnli} leads with 0.39 accuracy, macro F1 of 0.33, and weighted F1 of 0.32, but still trails full-evidence by ~10 points. \texttt{deberta-large-mnli} and \texttt{roberta-large-snli} perform worst, with 0.36 and 0.30 accuracy, respectively. Lower macro precision and recall suggest that span-based attribution yields incomplete evidence and reduced verification quality.
In CLIMATE-FEVER, span-based models also underperform. \texttt{roberta-large-snli} retains 0.64 accuracy, but its macro F1 drops to 0.32. \texttt{deberta-large-mnli} scores 0.58 accuracy and 0.51 weighted F1, yet its macro recall falls to 0.36. Overall, macro F1 ranges from 0.32 to 0.35, and weighted F1 stays around 0.50-0.54, confirming that span-based attribution sacrifices semantic depth for interpretability.

\subsection{Impact of \textit{threshold tuning} on DAVinCI with full evidence}

At a 0.7 threshold, DAVinCI achieves the highest overall accuracy. \texttt{deberta-large-mnli} leads with 0.47 accuracy and 0.60 macro precision. \texttt{roberta-large-mnli} (0.43) and \texttt{bart-large-mnli} (0.42) follow, while \texttt{roberta-large-snli} trails at 0.38 but retains strong precision (0.58). Raising the threshold to 0.8 slightly lowers accuracy (e.g., \texttt{deberta-large-mnli} to 0.46, \texttt{roberta-large-snli} to 0.34) and F1-scores, as more claims are filtered as ''Not Enough Info.'' Precision remains stable (0.58–0.60), but recall drops. At 0.9, DAVinCI becomes more conservative: accuracy dips (e.g., \texttt{deberta-large-mnli} to 0.45, \texttt{roberta-large-snli} to 0.31), and macro F1 declines (e.g., \texttt{roberta-large-snli to 0.19}). Precision stays high, but recall suffers, highlighting the trade-off between caution and coverage. Threshold 0.7 offers the best balance; 0.9 minimizes false positives. DAVinCI’s recalibration module enables flexible tuning for different risk profiles.

For CLIMATE-FEVER, DAVinCI performs strongly at 0.7. \texttt{roberta-large-snli} leads with 0.66 accuracy and 0.63 precision, though recall is lower (0.39). \texttt{deberta-large-mnli} and \texttt{roberta-large-mnli} reach 0.64 accuracy, with macro F1-scores of 0.40 and 0.43. \texttt{bart-large-mnli} follows at 0.62 accuracy and 0.40 F1. At 0.8, accuracy remains stable or improves (e.g., \texttt{roberta-large-mnli} to 0.65, \texttt{bart-large-mnli} to 0.63), and precision rises (e.g., \texttt{deberta-large-mnli} from 0.58 → 0.61). Recall dips slightly. At 0.9, accuracy holds or improves (e.g., \texttt{deberta-large-mnli to 0.65}), precision peaks (e.g., \texttt{roberta-large-snli} at 0.69), but F1-scores decline due to reduced recall. Accuracy trade-offs are minimal ($\leq$1.6\%), confirming DAVinCI’s ability to reduce false positives while maintaining strong performance.


\begin{table}[!htp]
\tiny \centering
\begin{tabular}{l|c|r|rrrrr}\toprule
\textbf{} &\textbf{Model} &\textbf{} &\textbf{P ↑} &\textbf{R ↑} &\textbf{F1 ↑} &\textbf{A ↑} \\\midrule
\multirow{12}{*}{\rotatebox{90}{\textbf{DAVinCI-full-evidence}}} &\multirow{3}{*}{\textbf{deberta-large-mnli}} &\textbf{} & \rowhighlight & \rowhighlight & \rowhighlight & \rowhighlight \textbf{0.48} \\
& &\textbf{macro avg} & \rowhighlight \textbf{0.57} & \rowhighlight \textbf{0.49} & \rowhighlight 0.41 & \rowhighlight \\
& &\textbf{weighted avg} & \rowhighlight \textbf{0.58} & \rowhighlight \textbf{0.48} & \rowhighlight 0.41 & \rowhighlight \\
&\multirow{3}{*}{\textbf{roberta-large-mnli}} &\textbf{} & & & &0.45 \\
& &\textbf{macro avg} &0.48 &0.46 &0.39 & \\
& &\textbf{weighted avg} &0.49 &0.45 &0.39 & \\
&\multirow{3}{*}{\textbf{bart-large-mnli}} &\textbf{} & \rowhighlight & \rowhighlight & \rowhighlight  & \rowhighlight 0.43 \\
& &\textbf{macro avg} & \rowhighlight 0.49 & \rowhighlight 0.44 & \rowhighlight 0.37 & \rowhighlight \\
& &\textbf{weighted avg} & \rowhighlight 0.49 & \rowhighlight 0.43 & \rowhighlight 0.37 & \rowhighlight \\
&\multirow{3}{*}{\textbf{roberta-large-snli}} &\textbf{} & & & &0.48 \\
& &\textbf{macro avg} &0.54 &0.48 & \textbf{0.48} & \\
& &\textbf{weighted avg} &0.56 & \textbf{0.48}  & \textbf{0.48} & \\ \midrule
\multirow{12}{*}{\rotatebox{90}{\textbf{DAVinCI-span-evidence}}} &\multirow{3}{*}{\textbf{deberta-large-mnli}} &\textbf{} & \rowhighlight & \rowhighlight & \rowhighlight & \rowhighlight 0.36 \\
& &\textbf{macro avg} & \rowhighlight 0.28 & \rowhighlight 0.38 & \rowhighlight 0.3 & \rowhighlight \\
& &\textbf{weighted avg} & \rowhighlight 0.28 & \rowhighlight 0.36 & \rowhighlight 0.29 & \rowhighlight \\
&\multirow{3}{*}{\textbf{roberta-large-mnli}} &\textbf{} & & & &0.33 \\
& &\textbf{macro avg} &0.35 &0.35 &0.3 & \\
& &\textbf{weighted avg} &0.35 &0.33 &0.3 & \\
&\multirow{3}{*}{\textbf{bart-large-mnli}} &\textbf{} & \rowhighlight & \rowhighlight & \rowhighlight & \rowhighlight 0.39 \\
& &\textbf{macro avg} & \rowhighlight 0.5 & \rowhighlight 0.4 & \rowhighlight 0.33 & \rowhighlight \\
& &\textbf{weighted avg} & \rowhighlight 0.51 & \rowhighlight 0.39 & \rowhighlight 0.32 & \rowhighlight \\
&\multirow{3}{*}{\textbf{roberta-large-snli}} &\textbf{} & & & &0.3 \\
& &\textbf{macro avg} &0.31 &0.33 &0.19 & \\
& &\textbf{weighted avg} &0.31 &0.3 &0.18 & \\
\bottomrule
\end{tabular}
\caption{Analysis of DAVinCI's results for full vs. span evidence in \textbf{FEVER dataset}. Numbers in bold show the best performance. (\textbf{P:} Precision, \textbf{R:} Recall, \textbf{F1:} F1-score, \textbf{A:} Accuracy)}\label{tab:fever-abl}
\end{table}


\begin{table}[!htp]
\tiny \centering
\begin{tabular}{l|c|r|rrrrr}\toprule
\textbf{} &\textbf{Model} &\textbf{} &\textbf{P ↑} &\textbf{R ↑} &\textbf{F1 ↑} &\textbf{A ↑} \\\midrule
\multirow{12}{*}{\rotatebox{90}{\textbf{DAVinCI-full-evidence}}} &\multirow{3}{*}{\textbf{deberta-large-mnli}} &\textbf{} & \rowhighlight & \rowhighlight & \rowhighlight & \rowhighlight 0.61 \\
& &\textbf{macro avg} & \rowhighlight 0.51 & \rowhighlight 0.43 & \rowhighlight 0.39 & \rowhighlight \\
& &\textbf{weighted avg} & \rowhighlight 0.6 & \rowhighlight 0.61 & \rowhighlight 0.54 & \rowhighlight\\
&\multirow{3}{*}{\textbf{roberta-large-mnli}} &\textbf{} & & & &0.61 \\
& &\textbf{macro avg} &0.49 & \textbf{0.47} &0.44 & \\
& &\textbf{weighted avg} &0.58 &0.61 &0.56 & \\
&\multirow{3}{*}{\textbf{bart-large-mnli}} &\textbf{} & \rowhighlight & \rowhighlight & \rowhighlight & \rowhighlight 0.58 \\
& &\textbf{macro avg} & \rowhighlight 0.49 & \rowhighlight 0.44 & \rowhighlight 0.4 & \rowhighlight \\
& &\textbf{weighted avg} & \rowhighlight 0.58 & \rowhighlight 0.58 &  \rowhighlight 0.54 & \rowhighlight \\
&\multirow{3}{*}{\textbf{roberta-large-snli}} &\textbf{} & & & & \textbf{0.65} \\
& &\textbf{macro avg} & \textbf{0.59} &0.4 &0.39 & \\
& &\textbf{weighted avg} & \textbf{0.64} & \textbf{0.65} &0.57 & \\ \midrule
\multirow{12}{*}{\rotatebox{90}{\textbf{DAVinCI-span-evidence}}} &\multirow{3}{*}{\textbf{deberta-large-mnli}} &\textbf{} & \rowhighlight & \rowhighlight & \rowhighlight & \rowhighlight 0.58 \\  
& &\textbf{macro avg} & \rowhighlight 0.51 & \rowhighlight 0.36 & \rowhighlight 0.32 & \rowhighlight \\
& &\textbf{weighted avg} & \rowhighlight 0.61 & \rowhighlight 0.58 & \rowhighlight 0.51 & \rowhighlight \\
&\multirow{3}{*}{\textbf{roberta-large-mnli}} &\textbf{} & & & &0.51 \\
& &\textbf{macro avg} &0.43 &0.36 &0.35 & \\
& &\textbf{weighted avg} &0.56 &0.51 &0.5 & \\
&\multirow{3}{*}{\textbf{bart-large-mnli}} &\textbf{} & \rowhighlight & \rowhighlight & \rowhighlight & \rowhighlight 0.54 \\
& &\textbf{macro avg} & \rowhighlight 0.48 & \rowhighlight 0.37 & \rowhighlight 0.33 & \rowhighlight \\
& &\textbf{weighted avg} & \rowhighlight 0.59 & \rowhighlight 0.54 & \rowhighlight 0.5 & \rowhighlight \\
&\multirow{3}{*}{\textbf{roberta-large-snli}} &\textbf{} & & & &0.64 \\
& &\textbf{macro avg} &0.45 &0.35 &0.32 & \\
& &\textbf{weighted avg} &0.57 &0.64 &0.54 & \\
\bottomrule
\end{tabular}
\caption{DAVinCI's performance comparison using full versus span evidence on the \textbf{CLIMATE-FEVER dataset}. Best results are highlighted in bold. (\textbf{P:} Precision, \textbf{R:} Recall, \textbf{F1:} F1-score, \textbf{A:} Accuracy)}\label{tab:climate-abl}
\end{table}


\begin{table}[!htp]
\tiny \centering
\begin{tabular}{l|c|r|rrrrr}\toprule
\textbf{} &\textbf{Model} &\textbf{} &\textbf{P ↑} &\textbf{R ↑} &\textbf{F1 ↑} &\textbf{A ↑} \\\midrule
\multirow{12}{*}{\rotatebox{90}{\parbox{2cm}{\textbf{DAVinCI-Recalibrated} \\ \textbf{(threshold = 0.7)}}}} &\multirow{3}{*}{\textbf{deberta-large-mnli}} &\textbf{} & \rowhighlight & \rowhighlight & \rowhighlight & \rowhighlight \textbf{0.47} \\
& &\textbf{macro avg} & \rowhighlight \textbf{0.6} & \rowhighlight \textbf{0.48} & \rowhighlight \textbf{0.4} & \rowhighlight \\
& &\textbf{weighted avg} & \rowhighlight \textbf{0.61} & \rowhighlight \textbf{0.47} & \rowhighlight \textbf{0.4} & \rowhighlight \\
&\multirow{3}{*}{\textbf{roberta-large-mnli}} &\textbf{} & & & &0.43 \\
& &\textbf{macro avg} &0.52 &0.45 &0.37 & \\
& &\textbf{weighted avg} &0.53 &0.43 &0.37 & \\
&\multirow{3}{*}{\textbf{bart-large-mnli}} &\textbf{} & \rowhighlight & \rowhighlight & \rowhighlight & \rowhighlight 0.42 \\
& &\textbf{macro avg} & \rowhighlight 0.51 & \rowhighlight 0.43 & \rowhighlight 0.36 & \rowhighlight \\
& &\textbf{weighted avg} & \rowhighlight 0.52 & \rowhighlight 0.42 & \rowhighlight 0.36 & \rowhighlight \\
&\multirow{3}{*}{\textbf{roberta-large-snli}} &\textbf{} & & & &0.38 \\
& &\textbf{macro avg} &0.58 &0.4 &0.33 & \\
& &\textbf{weighted avg} &0.59 &0.38 &0.32 & \\ \midrule
\multirow{12}{*}{\rotatebox{90}{\parbox{2cm}{\textbf{DAVinCI-Recalibrated} \\ \textbf{(threshold = 0.8)}}}} &\multirow{3}{*}{\textbf{deberta-large-mnli}} &\textbf{} & \rowhighlight & \rowhighlight & \rowhighlight & \rowhighlight 0.46 \\
& &\textbf{macro avg} & \rowhighlight 0.58 & \rowhighlight \textbf{0.48} & \rowhighlight 0.39 & \rowhighlight \\
& &\textbf{weighted avg} & \rowhighlight 0.6 & \rowhighlight 0.46 & \rowhighlight 0.39 & \rowhighlight \\
&\multirow{3}{*}{\textbf{roberta-large-mnli}} &\textbf{} & & & &0.43 \\
& &\textbf{macro avg} &0.53 &0.44 &0.36 & \\
& &\textbf{weighted avg} &0.54 &0.43 &0.36 & \\
&\multirow{3}{*}{\textbf{bart-large-mnli}} &\textbf{} & \rowhighlight & \rowhighlight & \rowhighlight & \rowhighlight 0.42 \\
& &\textbf{macro avg} & \rowhighlight 0.52 & \rowhighlight 0.43 & \rowhighlight 0.36 & \rowhighlight \\
& &\textbf{weighted avg} & \rowhighlight 0.54 & \rowhighlight 0.42 & \rowhighlight 0.36 & \rowhighlight \\
&\multirow{3}{*}{\textbf{roberta-large-snli}} &\textbf{} & & & &0.34 \\
& &\textbf{macro avg} &0.58 &0.37 &0.25 & \\
& &\textbf{weighted avg} &0.6 &0.34 &0.24 & \\ \midrule
\multirow{12}{*}{\rotatebox{90}{\parbox{2cm}{\textbf{DAVinCI-Recalibrated} \\ \textbf{(threshold = 0.9)}}}} &\multirow{3}{*}{\textbf{deberta-large-mnli}} &\textbf{} & \rowhighlight & \rowhighlight & \rowhighlight & \rowhighlight 0.45 \\
& &\textbf{macro avg} & \rowhighlight 0.58 & \rowhighlight 0.47 & \rowhighlight 0.38 & \rowhighlight \\
& &\textbf{weighted avg} & \rowhighlight 0.59 & \rowhighlight 0.45 & \rowhighlight 0.38 & \rowhighlight\\
&\multirow{3}{*}{\textbf{roberta-large-mnli}} &\textbf{} & & & &0.41 \\
& &\textbf{macro avg} &0.51 &0.44 &0.35 & \\
& &\textbf{weighted avg} &0.52 &0.41 &0.34 & \\
&\multirow{3}{*}{\textbf{bart-large-mnli}} &\textbf{} & \rowhighlight & \rowhighlight & \rowhighlight & \rowhighlight 0.4 \\
& &\textbf{macro avg} & \rowhighlight 0.55 & \rowhighlight 0.42 & \rowhighlight 0.34 & \rowhighlight \\
& &\textbf{weighted avg} & \rowhighlight 0.56 & \rowhighlight 0.4 & \rowhighlight 0.34 & \rowhighlight \\
&\multirow{3}{*}{\textbf{roberta-large-snli}} &\textbf{} & & & &0.31 \\
& &\textbf{macro avg} &0.56 &0.35 &0.19 & \\
& &\textbf{weighted avg} &0.57 &0.31 &0.17 & \\
\bottomrule
\end{tabular}
\caption{DAVinCI's performance on the \textbf{FEVER dataset} across three thresholds (0.7, 0.8, 0.9). Bolded values indicate the highest performance achieved. Threshold of \textbf{0.7} delivers the best results overall. (\textbf{P:} Precision, \textbf{R:} Recall, \textbf{F1:} F1-score, \textbf{A:} Accuracy)}\label{tab:fever-thresh}
\end{table}


\begin{table}[!htp]
\tiny \centering
\begin{tabular}{l|c|r|rrrrr}\toprule
\textbf{} &\textbf{Model} &\textbf{} &\textbf{P ↑} &\textbf{R ↑} &\textbf{F1 ↑} &\textbf{A ↑} \\\midrule
\multirow{12}{*}{\rotatebox{90}{\parbox{2cm}{\textbf{DAVinCI-Recalibrated} \\ \textbf{(threshold = 0.7)}}}} &\multirow{3}{*}{\textbf{deberta-large-mnli}} &\textbf{} & \rowhighlight & \rowhighlight & \rowhighlight & \rowhighlight 0.64 \\
& &\textbf{macro avg} & \rowhighlight 0.58 & \rowhighlight 0.43 & \rowhighlight 0.4 & \rowhighlight \\
& &\textbf{weighted avg} & \rowhighlight 0.64 & \rowhighlight 0.64 & \rowhighlight 0.56 & \rowhighlight \\
&\multirow{3}{*}{\textbf{roberta-large-mnli}} &\textbf{} & & & &0.64 \\
& &\textbf{macro avg} &0.57 & \textbf{0.46} & \textbf{0.43} & \\
& &\textbf{weighted avg} &0.63 &0.64 & \textbf{0.57} & \\
&\multirow{3}{*}{\textbf{bart-large-mnli}} &\textbf{} & \rowhighlight & \rowhighlight & \rowhighlight & \rowhighlight 0.62 \\
& &\textbf{macro avg} & \rowhighlight 0.54 & \rowhighlight 0.43 & \rowhighlight 0.4 & \rowhighlight \\
& &\textbf{weighted avg} & \rowhighlight 0.62 & \rowhighlight 0.62 & \rowhighlight 0.55 & \rowhighlight \\
&\multirow{3}{*}{\textbf{roberta-large-snli}} &\textbf{} & & & & \textbf{0.66} \\
& &\textbf{macro avg} &0.63 &0.39 &0.37 & \\
& &\textbf{weighted avg} &0.66 & \textbf{0.66} &0.56 & \\ \midrule
\multirow{12}{*}{\rotatebox{90}{\parbox{2cm}{\textbf{DAVinCI-Recalibrated} \\ \textbf{(threshold = 0.8)}}}} &\multirow{3}{*}{\textbf{deberta-large-mnli}} &\textbf{} & \rowhighlight & \rowhighlight & \rowhighlight & \rowhighlight 0.64 \\
& &\textbf{macro avg} & \rowhighlight 0.61 & \rowhighlight 0.41 & \rowhighlight 0.39 & \rowhighlight \\
& &\textbf{weighted avg} & \rowhighlight 0.66 & \rowhighlight 0.64 & \rowhighlight 0.55 & \rowhighlight \\
&\multirow{3}{*}{\textbf{roberta-large-mnli}} &\textbf{} & & & &0.65 \\
& &\textbf{macro avg} &0.58 &0.44 &0.42 & \\
& &\textbf{weighted avg} &0.64 &0.65 & \textbf{0.57} & \\
&\multirow{3}{*}{\textbf{bart-large-mnli}} &\textbf{} & \rowhighlight & \rowhighlight & \rowhighlight & \rowhighlight 0.63 \\
& &\textbf{macro avg} & \rowhighlight 0.57 & \rowhighlight 0.43 & \rowhighlight 0.4 & \rowhighlight \\
& &\textbf{weighted avg} & \rowhighlight 0.64 & \rowhighlight 0.63 & \rowhighlight 0.55 & \rowhighlight \\
&\multirow{3}{*}{\textbf{roberta-large-snli}} &\textbf{} & & & & \textbf{0.66} \\
& &\textbf{macro avg} &0.64 &0.38 &0.36 & \\
& &\textbf{weighted avg} &0.67 & \textbf{0.66} &0.55 & \\ \midrule
\multirow{12}{*}{\rotatebox{90}{\parbox{2cm}{\textbf{DAVinCI-Recalibrated} \\ \textbf{(threshold = 0.9)}}}} &\multirow{3}{*}{\textbf{deberta-large-mnli}} &\textbf{} & \rowhighlight & \rowhighlight & \rowhighlight & \rowhighlight 0.65 \\
& &\textbf{macro avg} & \rowhighlight 0.63 & \rowhighlight 0.39 & \rowhighlight 0.37 & \rowhighlight \\
& &\textbf{weighted avg} & \rowhighlight 0.67 & \rowhighlight 0.65 & \rowhighlight 0.55 & \rowhighlight \\
&\multirow{3}{*}{\textbf{roberta-large-mnli}} &\textbf{} & & & &0.65 \\
& &\textbf{macro avg} &0.62 &0.41 &0.39 & \\
& &\textbf{weighted avg} &0.66 &0.65 &0.56 & \\
&\multirow{3}{*}{\textbf{bart-large-mnli}} &\textbf{} & \rowhighlight & \rowhighlight & \rowhighlight & \rowhighlight 0.63 \\
& &\textbf{macro avg} & \rowhighlight 0.59 & \rowhighlight 0.41 & \rowhighlight 0.38 & \rowhighlight \\
& &\textbf{weighted avg} & \rowhighlight 0.65 & \rowhighlight 0.63 & \rowhighlight 0.55 & \rowhighlight \\
&\multirow{3}{*}{\textbf{roberta-large-snli}} &\textbf{} & & & & \textbf{0.66} \\
& &\textbf{macro avg} & \textbf{0.69}  &0.38 &0.35 & \\
& &\textbf{weighted avg} & \textbf{0.7} & \textbf{0.66} &0.55 & \\
\bottomrule
\end{tabular}
\caption{Evaluation of DAVinCI on the \textbf{CLIMATE-FEVER dataset} at thresholds 0.7, 0.8, and 0.9. The highest-performing results are highlighted in bold. Overall, \textbf{threshold=0.7} performs the best. (\textbf{P:} Precision, \textbf{R:} Recall, \textbf{F1:} F1-score, \textbf{A:} Accuracy)}\label{tab:climate-thresh}
\end{table}

\subsection{Ablation Insights}

For the FEVER dataset, full-evidence models outperform span-based ones by \textbf{9-18\%}, highlighting their superior performance. These models also deliver more stable macro and weighted F1-scores, reflecting improved class balance. All models benefit from richer context, with \texttt{roberta-large-snli} showing the largest F1 gain - from 0.19 to 0.48 - when switching to full evidence.
For the CLIMATE-FEVER dataset, across all models, DAVinCI with full-evidence consistently surpasses span-based versions, with accuracy improvements ranging from \textbf{+1.6\% to +19.6\%}. \texttt{roberta-large-mnli} sees the most significant jump \textbf{(~20\%)}, while \texttt{deberta-large-mnli} and \texttt{bart-large-mnli} improve by \textbf{5–7\%}. Even high-performing models like \texttt{roberta-large-snli} show further gains. These findings underscore the importance of comprehensive context in claim verification and demonstrate DAVinCI’s strength in leveraging full evidence for factual inference.
\textbf{Optimal Thresholding:} A threshold of \textbf{0.7} offers the best trade-off between precision and recall across datasets.
Together, these results affirm DAVinCI’s integrated approach to attribution and verification, showcasing its effectiveness in supporting reliable factual assessments.

\section{Conclusion and Future Work}

In this work, we introduced DAVinCI - a Dual Attribution and Verification framework designed to enhance the factual reliability of large language model outputs. By integrating evidence attribution and entailment-based verification into a unified pipeline, DAVinCI enables LLMs to not only generate claims but also justify and validate them in a transparent and auditable manner. Our experiments on FEVER and CLIMATE-FEVER demonstrate that DAVinCI consistently outperforms standard verification-only baselines. Through an ablation study, we show that both evidence quality and confidence calibration play critical roles in improving model reliability.

DAVinCI represents a step toward more trustworthy AI systems, especially in domains where factual correctness is essential. Future work will explore extensions to open-domain retrieval, multi-hop reasoning, and internal attribution tracing. We also envision applying DAVinCI to generative tasks, enabling LLMs to produce outputs that are not only fluent but also verifiable and source-aware. We also plan to expand its application to low-resource and multilingual environments. By bridging the gap between generation, attribution, and verification, DAVinCI contributes to the broader effort to make language models accountable.
\section{Ethics Statement}

This work does not involve the collection of new human data or personally identifiable information. All experiments were conducted using publicly available datasets (FEVER and CLIMATE-FEVER) that are widely used in the NLP community for factuality and claim verification research. We ensured that our framework, DAVinCI, operates transparently by attributing claims to evidence and calibrating confidence to reduce overconfident misclassifications. We release our code and evaluation scripts to promote reproducibility and responsible use. Our goal is to advance trustworthy NLP systems that support human oversight, not replace it.

\section{Discussion and Limitations}

Our experiments demonstrate that integrating attribution and verification into a unified pipeline significantly improves the factual reliability and interpretability of LLM outputs. DAVinCI consistently outperforms the baseline across multiple metrics. 

One of the key insights from our ablation study is the importance of evidence quality. Full evidence attribution yields more accurate and faithful verification than span-based extraction, which often loses critical context. This suggests that retrieval systems for LLMs should prioritize completeness and relevance over brevity. Additionally, our results show that confidence thresholds can be tuned to balance precision and recall, offering a simple yet effective mechanism for trust calibration.

Despite these strengths, DAVinCI has limitations. (i) It assumes access to high-quality evidence, which may not be available in open-domain settings. (ii) The verification module relies on static entailment models that may struggle with nuanced or multi-hop reasoning. (iii) Our current implementation does not model internal attribution (e.g., tracing claims to training data or prompts), which could further enhance interpretability. (iv) DAVinCI is evaluated only on English-language datasets (FEVER and CLIMATE-FEVER), which limits its applicability to multilingual or low-resource contexts. (v) Our confidence calibration relies on manually tuned thresholds that may not generalize across domains or tasks. 

Future work could address these limitations by integrating dense retrievers, multi-hop reasoning modules, and internal attribution techniques such as prompt tracing or activation clustering. We also envision extending DAVinCI to generative tasks, enabling LLMs to produce not only fluent text but also verifiable and source-aware outputs. Extending the framework to handle diverse linguistic structures and evidence formats remains an open challenge. Adaptive or learned calibration strategies could improve robustness and reduce the need for task-specific tuning. Human-in-the-loop evaluation could help assess the interpretability and trustworthiness of DAVinCI in real-world applications.

\section{Bibliographical References}\label{sec:reference}
\bibliographystyle{lrec2026-natbib}
\bibliography{lrec2026-example}

\end{document}